# Generative AI for Safe and Photorealistic Drone Light Shows

**Pascal Reinhold[1,*], Alexander Gräfe[1,*] and Sebastian Trimpe[1]**

[*]*Equal contributions*

[1]*Institute for Data Science in Mechanical Engineering, RWTH Aachen University*

Drone light shows are redefining aerial entertainment, yet their widespread adoption is bottlenecked by labor-intensive, manual animation. While generative AI promises an automated alternative, current frameworks fail to provide photorealism with fluid, dynamic motion. To address this limitation, we introduce SWAN, an end-to-end pipeline that synthesizes photorealistic, large-scale, and collision-free drone choreographies directly from text prompts. SWAN converts text into realistic reference videos and translates these pixel-space dynamics into physical swarm kinematics using a novel, adaptive point-tracking algorithm. Unlike existing trackers, this method maintains spatial coherence through severe occlusions and rapid topological shifts. A dedicated planner then allocates these trajectories to individual drones, while a subsequent safety filter ensures collision-free execution. We demonstrate scalability by safely orchestrating simulated 2,000-drone formations and validate physical feasibility on a dense real-world swarm of 49 quadcopters, operating everything entirely on standard consumer hardware. Combined, this work demonstrates how generative AI can be leveraged to automate multi-robot choreography design, providing an accessible new framework for drone light shows.

https://youtu.be/n3fEaqxEzho

https://github.com/Data-Science-in-Mechanical-Engineering/SWAN

## Introduction

Drone light shows have emerged as an important medium for large-scale public entertainment, increasingly replacing traditional pyrotechnics at major global events, such as New Year's celebrations, music festivals, and sporting events [1, 2, 3, 4, 5]. Beyond eliminating the air and noise pollution associated with fireworks [6, 7], drone swarms are able to render complex, dynamic figures in the night sky. This capability transitions aerial displays from the brief, explosive bursts of a firework into fluid, narrative-driven animations.

However, designing these choreographies remains a critical bottleneck. Current workflows rely on manual animation via general-purpose 3D software [8, 9] or specialized proprietary suites [10, 11, 12]. Customizing each performance to specific event themes through this manual pipeline is both costly and time-consuming. Consequently, while drone shows are used during high-budget events, the high cost of creating custom content hinders their widespread adoption and accessibility across the broader entertainment industry.

To overcome these bottlenecks, generative artificial intelligence (AI) offers a promising solution [13, 14, 15, 16, 17]. By using prompts to automatically generate drone shows, these models transition the labor-intensive manual process of drone show design to an automated, end-to-end pipeline. Consequently, this automation has the potential to drastically reduce cost and time. However, for such a framework to be viable in practice, it must satisfy three core demands of modern drone shows: **(i) high-fidelity photorealism**, **(ii) dynamic and physically realistic articulation**, and **(iii) scalability with guaranteed safety across thousands of agents**.

However, current generative frameworks fail to meet all these demands at once. ClipSwarm [13] iteratively refines formations using CLIP image-text similarity [18] to generate simple shapes from single words, while Gen-Swarms [14] leverages flow-matching 3D generative models to transform a word into static objects. FlockGPT [15] uses a Large Language Model (LLM) to generate 3D surfaces via code from sentence-level prompts. Hence, all three approaches synthesize realistic structures from text but yield formations that do not move. Conversely, SwarmGPT [16, 17] achieves dynamic choreographies via an LLM that outputs movement primitives (e.g., waves or spirals). However, this reliance on predefined primitives confines the approach to abstract geometric patterns, making it unable to display photorealistic subjects. Consequently, a framework meeting (i)-(iii) simultaneously does not currently exist.

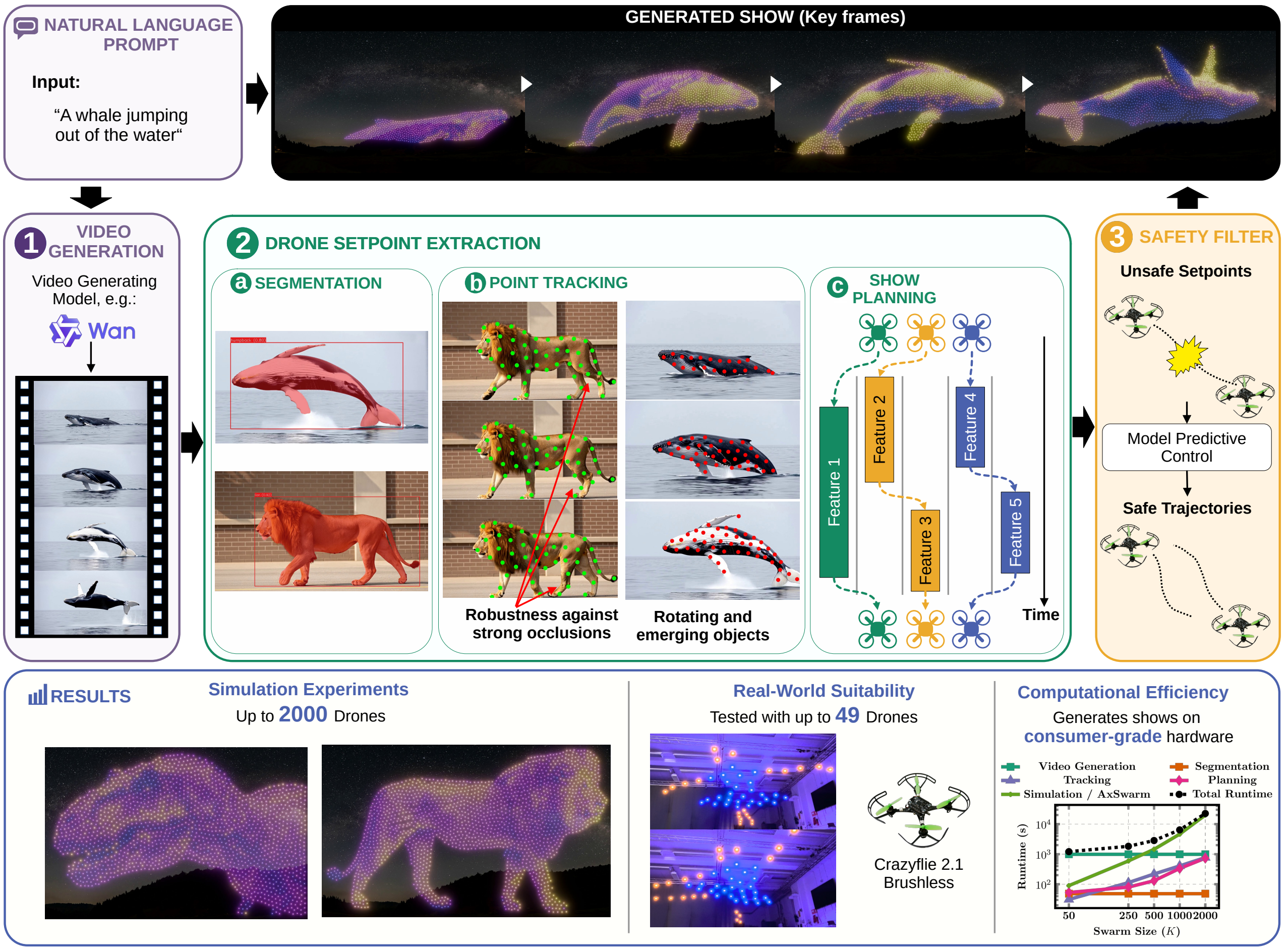

**Figure 1.** Overview of SWAN. The proposed framework translates a prompt into robot swarm trajectories through a three-stage pipeline. (1) Video Generation: A generative model generates a reference video based on the user's prompt. (2) Drone Setpoint Extraction: The primary subject is segmented from the video, and its visual features are tracked using an algorithm designed to handle severe occlusions and rotations. A planner then scales and assigns these extracted spatiotemporal feature trajectories to the drones. (3) Safety Filter: A model predictive controller processes the planned choreography to guarantee collision-free execution. In this paper, we demonstrate scalability with 2,000-drone simulations, deployment on 49 real-world quadcopters, and efficient end-to-end processing on standard desktop hardware.

To bridge this gap, we introduce *Swarm Animation from Natural Language* (SWAN) (Figure 1), the first end-to-end generative framework capable of synthesizing photorealistic, dynamic, large-scale, and safe drone shows directly from text prompts. The main idea of SWAN is to achieve this photorealism through video generation. It converts text prompts into photorealistic reference videos via state-of-the-art models like Wan 2.2 [19]. SWAN then translates these videos into drone shows by isolating the primary subject within this generated video and extracting feature trajectory segments to serve as swarm setpoints. Finally, the framework assigns these segments to individual drones to assemble a continuous choreography via a dedicated planner module. This choreography is subsequently processed by a safety filter (AxSwarm [16, 17, 20]) to ensure kinematically feasible and collision-free execution.

The primary challenge in this pipeline is the point-tracking stage. Existing point-tracking algorithms, like CoTracker [21, 22], typically assume rigid motion and simple occlusion models, leading to failure under the strong deformations and complex dynamics inherent to the generated videos. SWAN overcomes this challenge through a novel, adaptive tracking mechanism. Rather than strictly assigning trajectories to persistent visual features, our method adapts to the evolving scene topology. Our system continuously monitors feature density and visibility. It terminates trajectories for permanently occluded elements and autonomously initializes tracking for newly emerging ones. This adaptability ensures that we can extract point trajectories on the entire object even during severe occlusions and topological shifts.

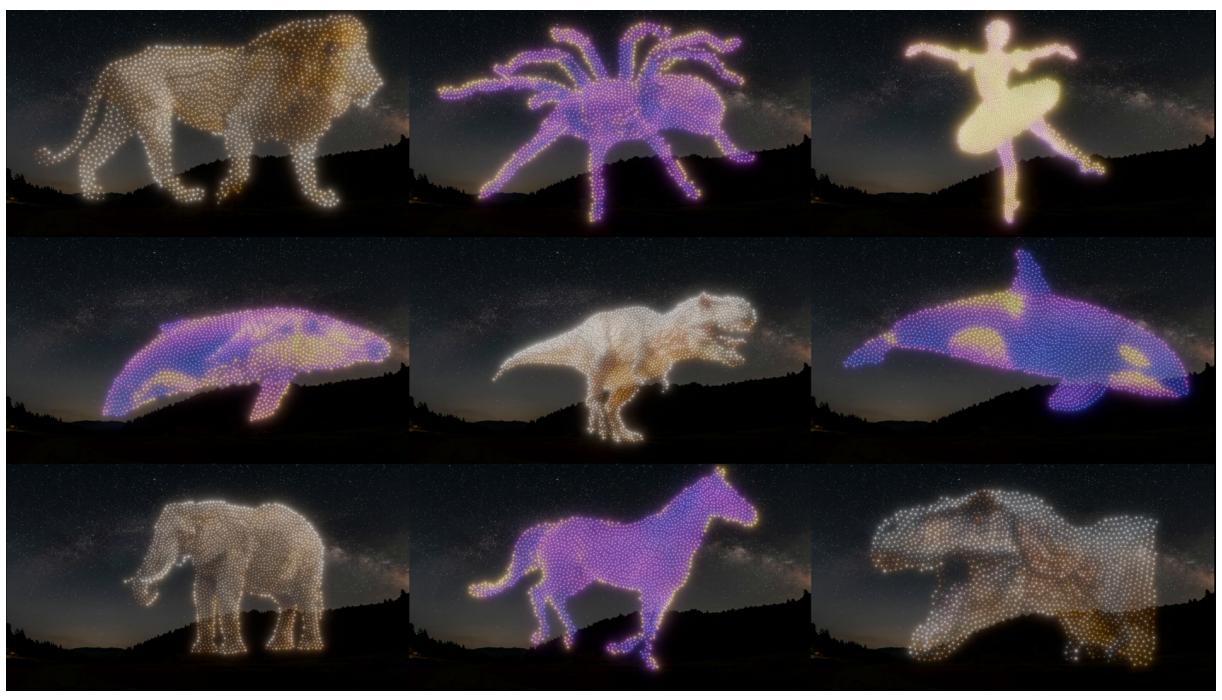

**Movie 1.** Overview of SWAN. This video demonstrates the generated shows executed in both simulation and reality, corresponding to the results in Figures 2 and 4. The thumbnail shows a collage of generated drone shows. The video can be found under `http://tiny.cc/SWANGenerativeAI`.

The second challenge is to translate these fragmented, pixel-space trajectories into swarm setpoint trajectories. To resolve this, a planner module transforms these segments into the physical space, scaling them to respect the drones' physical dimensions, velocity bounds, and acceleration limits. The system then optimally allocates these scaled trajectories to individual drones within the swarm. Finally, the planner synthesizes kinematically viable transitions between the segments and generates continuous setpoints that can be passed to the safety filter.

In this work, we show that this pipeline can translate natural language prompts into photorealistic drone shows. We first demonstrate its scalability by generating choreographies, such as a roaring Tyrannosaurus rex and a breaching whale, for up to 2,000 drones in simulation (Movie 1, Figure 2). Furthermore, we prove the system's real-world viability by deploying generated trajectories onto a swarm of 49 Crazyflie quadcopters (Movie 1), achieving the first fully AI-generated physical drone show of this complexity. Importantly, the entire pipeline remains computationally lightweight and runs entirely on a standard consumer-grade desktop computer (AMD Ryzen 5 3600, 64 GB RAM, and Nvidia RTX 4060 Ti with 16 GB VRAM).

## Results

We demonstrate SWAN's ability to translate natural language directly into safe, physically deployable drone shows through a comprehensive evaluation. We first show its end-to-end performance and scalability via multi-agent simulations. Afterwards, we evaluate the physical feasibility through real-world hardware deployments. Given that one of SWAN's main technical challenges is the setpoint extraction, we subsequently benchmark our novel point-tracking module against the current state-of-the-art. Finally, this section assesses SWAN's computational efficiency and scalability with the swarm size. Our evaluation yields five key findings:

1. **Text-to-choreography generation:** SWAN synthesizes dynamic, photorealistic, and collision-free drone shows directly from natural language prompts, eliminating the need for manual animation.
2. **Scalability:** The framework orchestrates complex, densely packed formations of 2,000 drones in simulation.
3. **Real-world suitability:** SWAN generates physically viable and safe trajectories that are suitable for real-world control, enabling real-world deployment on a swarm of 49 Crazyflie quadcopters.
4. **Tracking under strong occlusion:** SWAN's point tracker maintains coherence through frequent visual occlusions and complex rotational dynamics, scenarios where the current state-of-the-art method, CoTracker3 [22], fails.
5. **Computational efficiency:** Despite the complexity of coordinating thousands of drones, the end-to-end pipeline operates efficiently on standard consumer-grade hardware, generating 2,000-drone shows in a few hours.

### End-to-end Analysis: Simulation

To prove that SWAN can safely and reliably create photorealistic, scalable drone shows, we first test the entire system in simulation. Our simulation uses the first-principles dynamics model of Crazyflie quadcopters using Crazyflow [23], whose parallel simulation on GPUs allows for fast simulation.

Figure 2 and Movie 1 illustrate four choreographies, a Tyrannosaurus rex, a lion, an elephant, and a whale, generated from basic text prompts. These drone shows achieve high visual fidelity and capture details like fur, teeth, and tongues. Moreover, it achieves realistic movements, also capturing the subtle dynamics of subsurface muscle movement, extending well beyond simple rigid articulation.

Achieving these detailed drone shows requires tailored fleet management and a dynamic allocation strategy. Rather than permanently anchoring drones to individual tracking points, SWAN continuously assigns resting drones to newly emerging visual features and reallocates those whose targets become occluded via the combination of point tracker and planner module. While the top of Figure 2 simulates physical deployment by illuminating only actively engaged drones, its last row exposes this underlying swarm mechanic by visualizing the complete swarm during the whale sequence. Drones initialize in a grid (e.g., grounded charging pods) and take off to track their assigned features across the primary image plane. As the whale breaches the water, the system allocates new drones to emerging features until nearly the entire fleet is engaged. During the rotation of the whale, drones are reassigned from disappearing features to new features. As can be seen in Movie 1, after flying the show, the drones then return to the starting positions for landing. This demonstrates that our pipeline covers the entire show operation: from start, through the actual show, to landing.

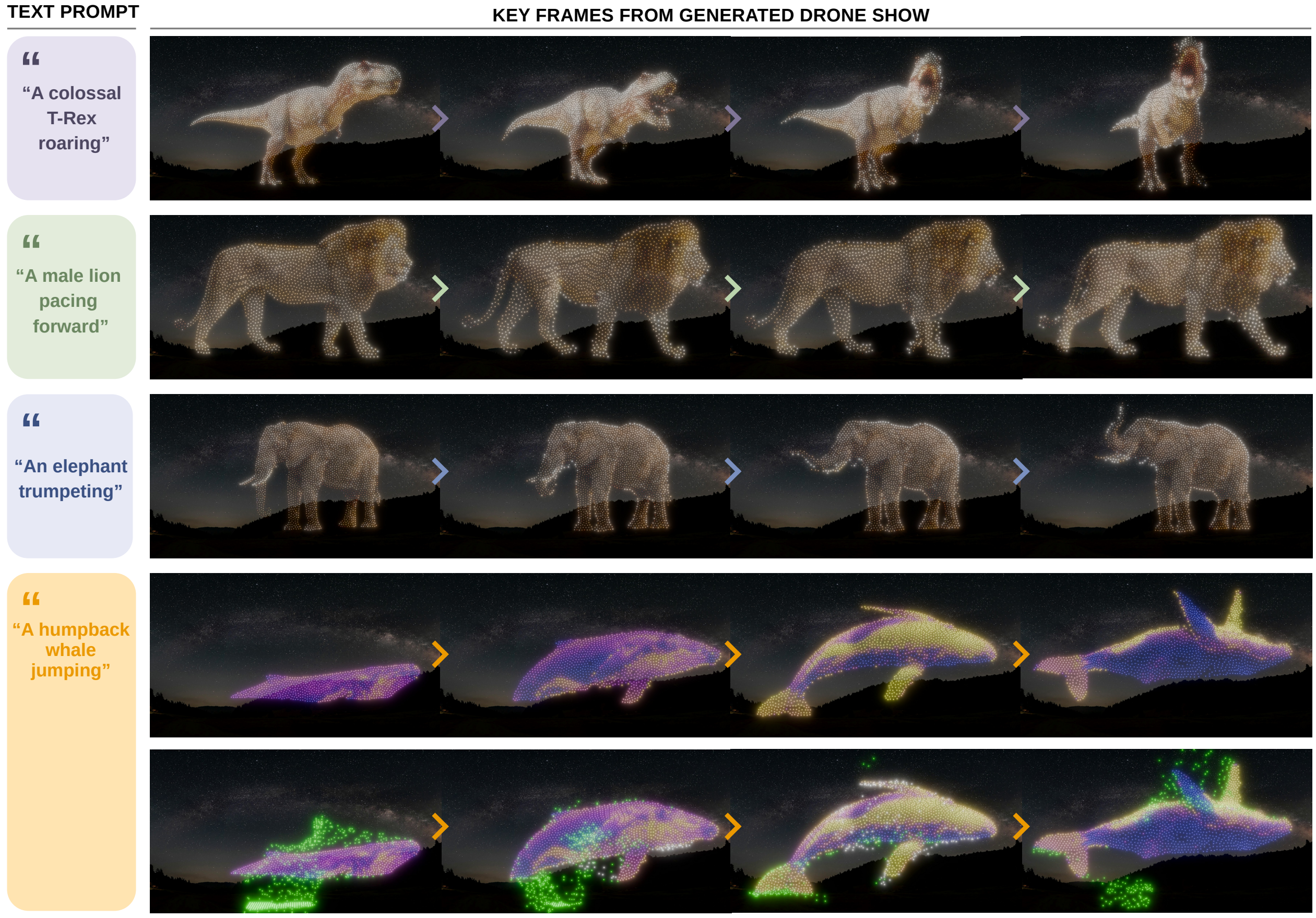

**Figure 2.** Visualization of selected choreographies. All shows are flown in simulation with 2,000 drones. Each choreography is labeled with a shortened version of its generation prompt. All show life-like movement of the rendered animals, demonstrating that SWAN can generate photorealistic drone shows from prompts. The bottom row shows the whale formation with LEDs on all drones turned on, with non-allocated drones' LEDs turned green in case they are newly allocated and white in case they are reassigned to a new feature.

Importantly, the show's complexity does not compromise operational safety. Figure 3 tracks the minimum inter-agent distances across all four generated shows. The swarm adheres to the defined safety distances, resulting in collision-free operations.

Ultimately, these end-to-end simulation results demonstrate that SWAN can translate text into realistic, large-scale, and safe drone shows.

## End-to-end Analysis: Real-World Swarm

We next validate the physical feasibility of these generated choreographies through real-world hardware deployments. To this end, we deployed a swarm of up to 49 Crazyflie Brushless quadcopters within a confined $7\,\text{m} \times 7\,\text{m} \times 2\,\text{m}$ flight volume.[1]

The physical setup utilized a Lighthouse system [24] for precise indoor positioning and the Crazyflie Python Library V2 [25] for swarm control. The quadcopters are equipped with LEDs at their bottom [26]. We translate the trajectories generated by SWAN into Bézier splines and transmit them to individual drones via radio. Once airborne, the quadcopters track these trajectories while continuously broadcasting their internal state estimates back to the base station at 20 Hz for logging.

We evaluate the system's real world performance across three distinct choreographies: a karate fighter, a ballerina, and a hand signing "I love you" (ILY) in American Sign Language. While the first two shows were generated from synthetic video from a generative AI model (Wan 2.2 [19]), the ASL sequence was extracted from a real-world video. With this, we demonstrate that SWAN can also utilize real videos as source material.

As demonstrated in Figure 4 and Movie 1, SWAN enables a real swarm to execute drone shows directly from a single prompt. The physical flight closely mirrors the underlying source video, with the swarm maintaining collision avoidance throughout the choreography.

[1]We gratefully acknowledge the Bitcraze team for hosting these experiments at their facility.

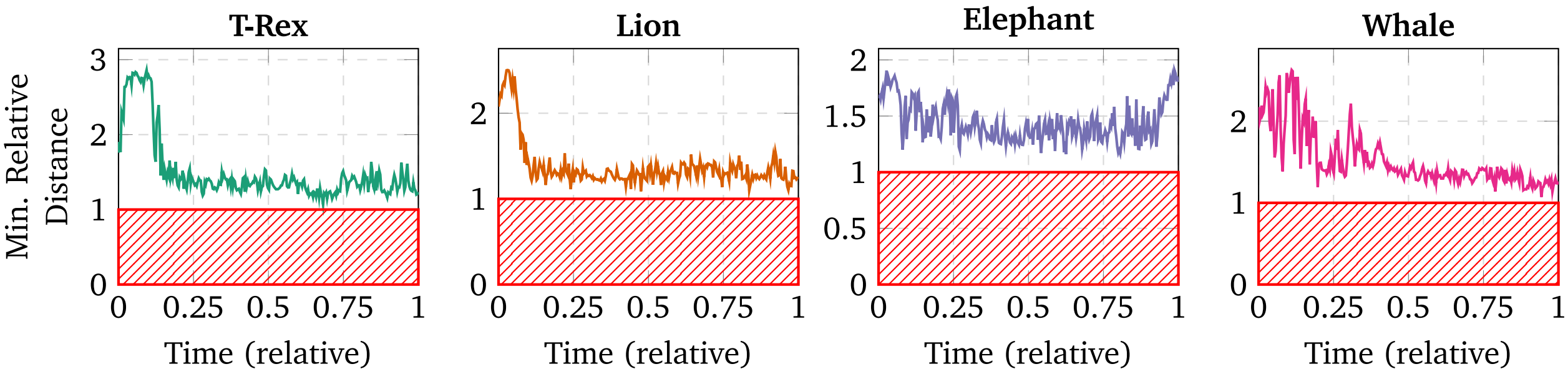

**Figure 3.** Minimum inter-drone distance over time for each choreography sequence shown in Figure 2. For ease of presentation, we show the relative time and distances, normalizing by the show length and minimum allowable distance set in the safety filter. Over time, the minimum distance between drones is almost always close to the minimum allowable distance but does not cross it.

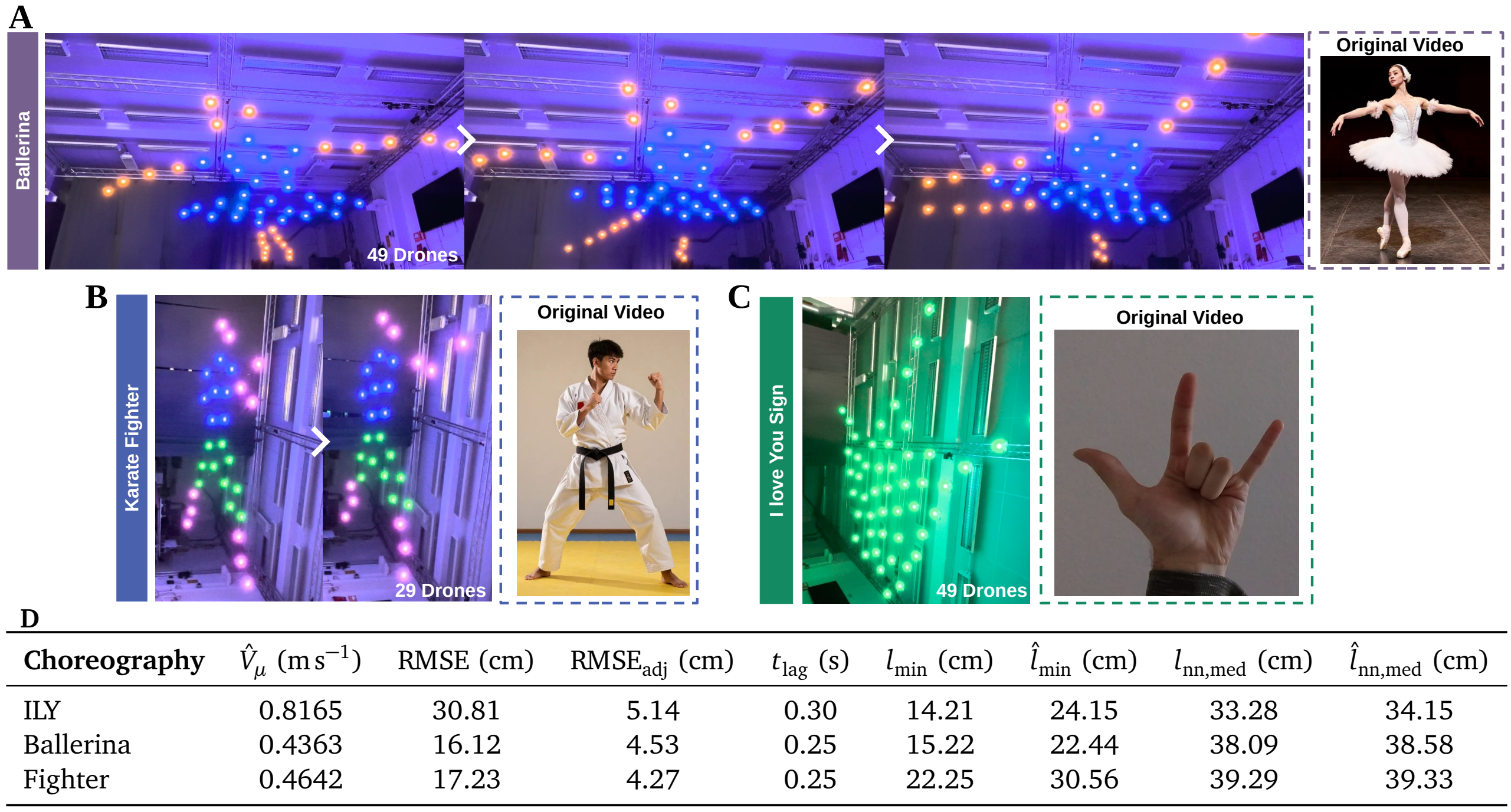

| Choreography | $\hat{V}_\mu$ ($\mathrm{m\,s^{-1}}$) | RMSE (cm) | $\mathrm{RMSE_{adj}}$ (cm) | $t_{lag}$ (s) | $l_{min}$ (cm) | $\hat{l}_{min}$ (cm) | $l_{nn,med}$ (cm) | $\hat{l}_{nn,med}$ (cm) |
|---|---|---|---|---|---|---|---|---|
| ILY | 0.8165 | 30.81 | 5.14 | 0.30 | 14.21 | 24.15 | 33.28 | 34.15 |
| Ballerina | 0.4363 | 16.12 | 4.53 | 0.25 | 15.22 | 22.44 | 38.09 | 38.58 |
| Fighter | 0.4642 | 17.23 | 4.27 | 0.25 | 22.25 | 30.56 | 39.29 | 39.33 |

**Figure 4. Top:** Generated choreographies featured in our real-world experiments. We perform hardware experiments with three different shows: a ballerina (**A**), a karate fighter (**B**), and a hand making an "I love you" (ILY) sign (**C**). Figures (**B**) and (**C**) are rotated for better visibility. **D** Reference velocities $\hat{V}_\mu$, tracking error, lag-adjusted tracking error, estimated lag for reference, minimum distances for flown ($l_{min}$) and reference trajectories ($\hat{l}_{min}$), alongside median nearest-neighbor distances for flown ($l_{nn,med}$) and reference ($\hat{l}_{nn,med}$) trajectories.

This safety results from precise trajectory tracking. While our analysis (Figure 4 **D**) reveals a systematic phase lag of approximately 300 ms, an expected artifact of radio communication delays and internal control loops, compensating for this temporal shift yields a mean positional error of just 5 cm. This matches the baseline tracking performance reported for AxSwarm [20].

Crucially, this tracking accuracy directly translates to operational safety. By strictly adhering to their generated trajectories, the quadcopters avoided all inter-agent collisions. Furthermore, an analysis of inter-drone distances (Figure 4 **D**) confirms a median separation of more than 30 cm. Although the minimum recorded distance dropped to 14 cm, a value smaller than the physical footprint of the quadcopters, no actual collisions occurred during the experiments. We therefore attribute this discrepancy to estimation errors within the drones' Kalman filters.

Ultimately, these hardware experiments demonstrate that the collision avoidance demonstrated in our simulations also holds true in reality and that SWAN can generate trajectories suitable for real drones.

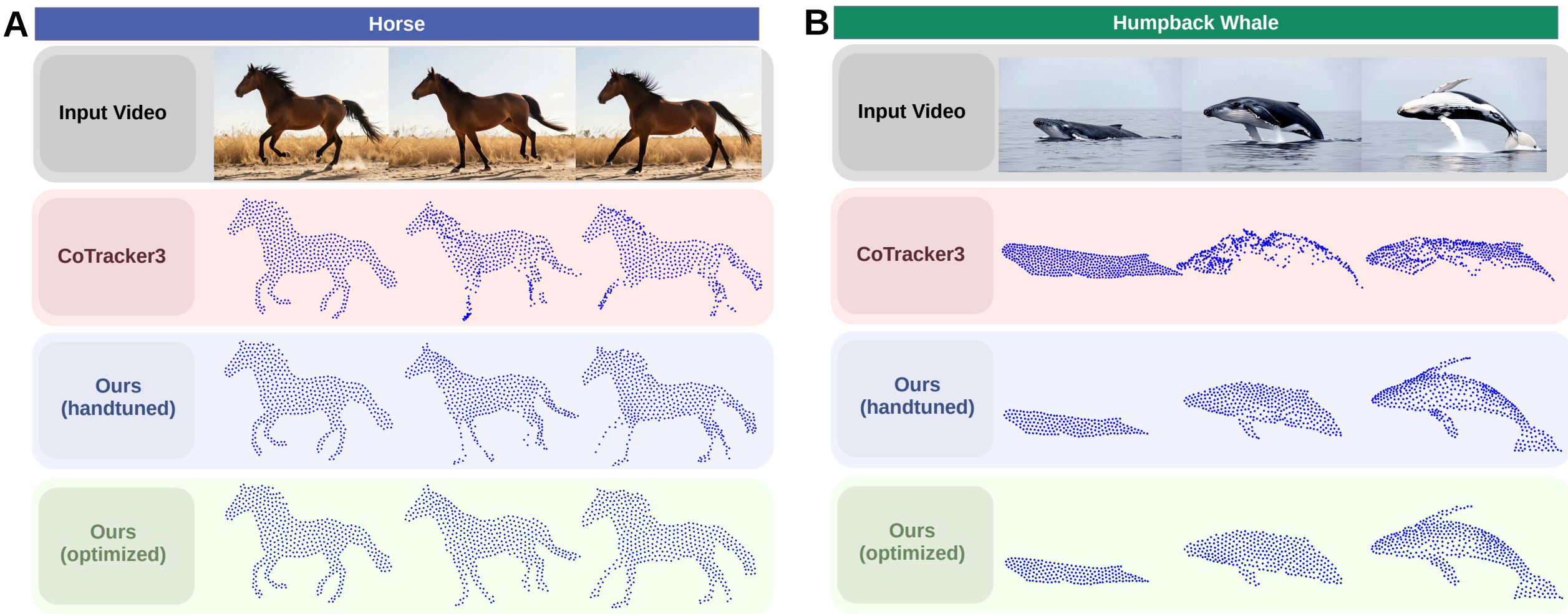

**Figure 5.** Comparison of CoTracker3 and our point tracking method across two example videos. **A**: CoTracker3 fails to track the limbs of running animals due to frequent occlusions. **B**: CoTracker3 cannot extract points from rotating and emerging objects, while our method successfully tracks them.

## Analysis of Point Tracking Method

Having demonstrated the end-to-end scalability of SWAN in simulations and physical deployments, we now analyze its core algorithmic driver: the point-tracking module. To contextualize the performance, we benchmark our approach against CoTracker3 [22], the current state-of-the-art in point tracking. Furthermore, to evaluate our design choices, we ablate our method using a hand-tuned configuration and a variant optimized via Optuna [27]. The latter leverages multi-objective Bayesian optimization [28] to maximize spatial coverage while minimizing drone churn rate (i.e., the frequency of required reassignments).

Figure 5 illustrates the comparison across two sequences that represent common challenges in typical source videos. The first, a running horse, features frequent self-occlusions of its limbs. The second, a breaching whale, features an object that both emerges and rotates, where new areas on its body continuously appear and disappear.

As demonstrated by the running horse sequence, standard methods degrade under articulated self-occlusions. Here, CoTracker3 shows tracking drift: points jump across intersecting limbs, leaving entire anatomical structures untracked and fundamentally breaking the formation's physical appearance. We hypothesize that this occurs due to a lack of similar data during CoTracker3's training and because the legs are of similar color and structure, which makes it difficult for CoTracker3 to differentiate them. Conversely, our method maintains strict spatial coherence across the occluding limbs. While the hand-tuned variant struggles to fully cover some limbs, yielding an incomplete visual depiction, the optimized variant is capable of fully tracking the horse's gait.

The limitations of current point trackers are most apparent in the final formation: a breaching humpback whale. Here, CoTracker3 fails entirely, losing structural coherence and causing the formation to collapse. In contrast, the proposed tracker handles these dynamics by continuously tracking new features and terminating the tracking of disappearing ones. Through this, it is capable of extracting swarm setpoints capable of representing the entire jumping whale. Notably, while both of our variants succeed here, the optimized configuration enforces a more uniform drone distribution across the whale's surface area.

Ultimately, these results demonstrate that our tracking algorithm significantly surpasses the state-of-the-art in handling the severe occlusions, rotations, and topological shifts inherent to complex animations. This robustness enables SWAN to translate videos into drone shows.

## Computational Performance Analysis

To evaluate the computational requirements of our method, we measured the runtime of each algorithmic stage (video generation, segmentation, point tracking, trajectory generation, and safety filter) across varying swarm sizes. The evaluation was conducted on a consumer-grade desktop computer (AMD Ryzen 5 3600, 64 GB RAM, and Nvidia RTX 4060 Ti with 16 GB VRAM).

As shown in Figure 6, the total pipeline duration scales from 6.2 min for 50 drones to 6.2 h for 2000 drones. This increase is primarily caused by the safety filter. For 50 drones, the filter accounts for 7.4 % (1.5 min) of the total runtime, whereas for 2000 drones, it consumes 88.7 % (332.3 min). This increase results from two factors. First, the growing number of drones creates larger optimization problems. Second, larger swarms require physically larger formations, which de-mand longer flight distances and consequently extend the overall duration of the show.

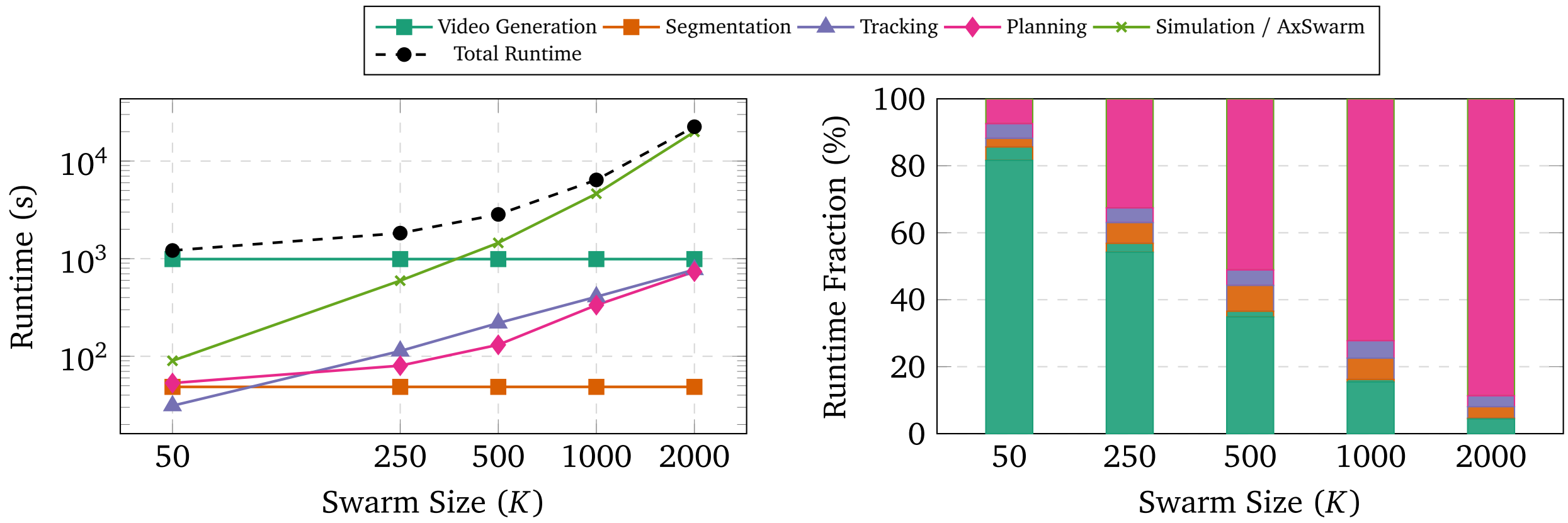

**Figure 6.** Pipeline runtime analysis. The upper plot shows mean total and stagewise runtimes across swarm sizes. The lower shows fractions of total runtime corresponding to each stage. The more drones there are in the swarm, the longer the tracking, trajectory generation, and safety filter take. The safety filter dominates the duration, taking almost the entire duration for 2000 drones.

In contrast, show generation without the safety filter remains efficient, taking 18.7 min for 50 drones and 24.44 min for 2000 drones. Because the input video and segmentation remain identical regardless of swarm size, the video generation and segmentation stages are constant at approximately 16.5 min and 48.59 s, respectively. Point tracking duration increases from 0.5 min to 12.9 min, while trajectory generation scales from 0.9 min to 12.26 min. Notably, both stages exhibit sublinear scaling relative to swarm size. A 40-fold increase in the number of drones only results in a 24.8-fold and 13.8-fold increase in computation time, respectively.

These results demonstrate that, depending on the swarm size, SWAN can generate shows on standard hardware in a timeframe ranging from a few minutes to a few hours. Because the initial stages execute in under half an hour, we recommend a two-step workflow in practice. Users can generate and preview the unconstrained formation, iteratively adjust the prompt as needed, and only execute the computationally expensive safety filter once the visual design is finalized.

Furthermore, we note that the hardware used in these experiments represents the lower end of desktop-grade specifications. Upgrading the desktop GPU would likely reduce the reported computation times. Interestingly, we observed that server-grade hardware, e.g., H100 GPUs, actually led to longer runtimes for the safety filter. We presume this is caused by the absence of dedicated raytracing cores, which are necessary to efficiently handle the branching logic within the safety filter.

## Discussion

The design of drone shows is currently bottlenecked by labor-intensive, manual design. In this work, we introduced SWAN, an end-to-end generative framework that bridges the gap between natural language prompts and large-scale, physically deployable aerial choreographies. By safely coordinating formations of up to 2,000 agents in simulation and validating these trajectories on 49 physical quadcopters, we have demonstrated that our framework efficiently translates text into realistic, dynamic drone shows, removing the need for manual animation.

The proposed pipeline operates in three distinct stages: text-to-video generation, setpoint extraction, and safety filtering. Our experiments validated the efficacy of each stage. First, the video generation module successfully produces photorealistic videos suitable for downstream processing. From these videos, the setpoint extraction module isolates feature trajectories on the primary object to serve as drone setpoints. We demonstrated that the adaptive point-tracking mechanism within this module is a core enabler of the pipeline. It significantly outperforms state-of-the-art methods like CoTracker3 [22] by robustly handling the deformations, self-occlusions, and topological shifts typical of synthesized videos. By dynamically managing feature life-cycles, SWAN successfully extracts the spatiotemporal trajectories required to render dynamic, lifelike subjects, ranging from a roaring Tyrannosaurus rex to a human ballerina, without requiring manual animation. Finally, our results confirmed that the safety filter guarantees collision-free navigation without compromising the structural integrity of the overall show. Crucially, SWAN is highly resource-efficient and can be executed entirely on a consumer-grade desktop computer, bypassing the need for a high-performance compute cluster.

Despite these advancements, SWAN also has some limitations, which we want to discuss:

- **Refining Intersections:** While our algorithm robustly tracks intersecting limbs without losing

them, the exact crossing moment can currently produce minor visual artifacts. These artifacts are caused by the complex interplay between the safety filter and the underlying tracking mechanism. Smoothing this interaction remains a target for future algorithmic refinement.
- **Dimensionality of Projections:** Currently, the generated shows are 2D projections derived from the source video's visual plane. Expanding these choreographies into true 3D animations would significantly enhance visual impact and realism. This could be achieved by integrating depth estimators to extrapolate 3D structural data from 2D videos, or by incorporating emerging text-to-3D animation models [29] directly into the SWAN pipeline.
- **Unstructured Objects:** Because SWAN relies on the structural coherence of a segmented object, the method is unsuited for chaotic or highly fragmented visual effects. For phenomena lacking a cohesive primary object, such as fireworks, scattering debris, or splashing water, the underlying algorithms cannot extract reliable, continuous feature trajectories.

# Methods

This section details the proposed SWAN framework, describing each stage outlined in Figure 1, namely video generation, segmentation, point tracking, show planning, and safety filtering. An implementation of our method can be found at https://github.com/Data-Science-in-Mechanical-Engineering/SWAN.

## Video Generation

This stage has three main objectives: (1) generating a video that accurately reflects the user prompt, (2) optimizing the visual output for downstream segmentation and point tracking, and (3) extracting a corresponding segmentation prompt for the segmentation stage.

**Prompt Expansion** We expand the initial user prompt into a structured format containing all parameters necessary for video generation to mimic the structured captions used during the training of video generating models [30]. Specifically, an LLM (Qwen 3.5-9B [31]) converts the user prompt into JSON data detailing lighting, camera angle, shot size, background, subject appearance (including initial pose), and subject motion. To ensure the resulting videos are optimal for downstream visual processing, the LLM supplements missing information based on the following constraints:

- **Diffuse lighting** to minimize shadows and reflections.
- **Optimal framing** to keep the main subject fully visible and prominently sized, utilizing tracking or panning shots for dynamic subjects.
- **Orthogonal side views** to produce distinctive silhouettes.
- **Static, uncluttered backgrounds** (e.g., avoiding tall grass) to maximize contrast with the subject.

From the generated JSON, we derive two text prompts for the subsequent generation stage: (a) an image prompt concatenating the lighting, subject, and camera details, and (b) a video prompt appending the motion description to the image prompt. Finally, it also generates a segmentation prompt for the segmentation stage.

**Video Generation Model** Our framework supports two generative pipelines: direct Text-to-Video (T2V) and Text-to-Image-to-Video (T2I2V) [19]. In the T2V workflow, a model synthesizes the video sequence entirely from the video prompt. In contrast, the T2I2V workflow is a two-step process. First, a text-to-image model generates an initial start frame from the image prompt to establish the environment and initial pose. Second, an image-to-video model renders the final sequence using both this generated start frame and the video prompt.

At the time of writing, we found that the T2I2V workflow utilizing Z-Image [32] for text-to-image generation and Wan 2.2 [19] for video generation yields the best results among openly available models. Consequently, all videos presented in this paper were produced using this approach. However, given the current evolution of generative video models, our framework is designed to accommodate future architectures that may offer better performance.

## Segmentation

The first step of our setpoint extraction approach is video segmentation. We input the generated video and an LLM-derived segmentation prompt, obtained during the video generation stage, into a segmentation model, such as SAM 3 [33]. This produces a binary mask across all frames that isolates the target object. Within our pipeline, this mask serves as the tracking ground truth, defining the spatial boundaries of the tracking points and enabling outlier detection.

## Point Tracking

The point tracking module translates the segmented object into drone trajectories. To handle newly emerging and vanishing features, we track points over short horizons rather than extracting full-length trajectories. This yields a set of discrete trajectory segments $\mathcal{S} = \{s_1, s_2, \dots, s_N\}$, defined as

$$s_i = \left(x_i^{(t)}\right)_{t_{\text{start},i}}^{t_{\text{end},i}}, \quad x_i^{(t)} \in \mathbb{R}^2, \tag{1}$$

where $t_{\text{start},i}$ and $t_{\text{end},i}$ represent the start and end times, and $x_i^{(t)}$ is the 2D tracked point at time $t$. Consequently, at any given time $t$, the active segment indices and their corresponding tracked points are formulated as

$$\mathcal{P}^{(t)} = \left\{ i \in \{1, \ldots, N\} \mid t \in \{t_{\text{start},i}, \ldots, t_{\text{end},i}\} \right\} \quad (2)$$

$$\mathbf{X}^{(t)} = \left\{ x_i^{(t)} \mid i \in \mathcal{P}^{(t)} \right\}. \quad (3)$$

The number of concurrently active segments, $|\mathcal{P}^{(t)}|$, is bounded by a maximum active capacity $\hat{K}_{\text{max}} \leq K$, where $K$ is the total swarm size. If $\hat{K}_{\text{max}}$ is chosen too large, the later planner module might not find a valid solution. We present our strategy to choose $\hat{K}_{\text{max}}$ in the corresponding planner section. To maintain uniform drone density, SWAN scales the target number of active segments $\hat{K}^{(t)}$ at time $t$ proportionally with the mask's current area $A^{(t)}$ relative to its maximum area over the video, $A_{\text{max}}$

$$\hat{K}^{(t)} = \left\lceil \frac{\hat{K}_{\text{max}}}{A_{\text{max}}} A^{(t)} \right\rceil. \quad (4)$$

Algorithm 1 details the extraction of these trajectory segments. We first provide a high-level overview of the algorithm before detailing specific components below.

Rather than processing the entire video in a single continuous pass, our framework employs sequential tracking intervals interleaved with intermediate corrections. Following initialization at $t = 0$, the video is processed via two nested loops. The outer loop advances by $h$ frames per iteration (line 3), utilizing a pre-trained general-purpose point tracker, $\psi_{\text{T}}$, initialized with $\mathbf{X}^{(t)}$, to predict raw trajectories. In our implementation, we use CoTracker3 [22]. These predictions are subsequently refined frame-by-frame within the inner loop (line 5). Instead of directly assigning points to destinations the pre-trained point tracker predicted, this loop models them as inertia-free particles within a force field. This field guides the tracked points along their predicted destinations while enforcing collision avoidance and segmentation mask boundaries (line 6). Any points exceeding a local density threshold or drifting outside the mask are removed. Their corresponding trajectory segments are then terminated and appended to $\mathcal{S}$ (lines 7–12). At the conclusion of each outer iteration, new points are injected into low-coverage regions of the mask until the target capacity $\hat{K}^{(t)}$ is reached.

The remainder of this section details the initialization, force field modeling, detection of overcrowded point clusters and outliers, and the point insertion mechanism. All hyperparameters for the pipeline are optimized via Bayesian optimization as outlined in the comparative analysis of the point trackers.

**Initialization.** We initialize tracking by uniformly distributing $\hat{K}^{(1)}$ points within the initial segmentation mask. To ensure a regular spatial distribution, we apply $K$-means clustering with $\hat{K}^{(1)}$ clusters to the mask's interior pixels and assign the resulting centroids as the initial points, closely matching centroidal Voronoi tesselation (CVT) [34]. To parameterize subsequent distance-based interactions (e.g., collision detection), we establish a reference packing distance, $l_\rho$. This is determined by applying the same CVT approximation using $\hat{K}_{\text{max}}$ points on the frame containing the largest segmentation mask, defining $l_\rho$ as the average minimum pairwise distance among these sampled centroids.

**Force Field Modeling.** Rather than employing direct tracking, we model the extracted points as particles in an overdamped system. Because inertia is negligible in such systems, a particle's velocity is directly proportional to the applied forces [35]. We define the velocity $\dot{\mathbf{x}}_i$ of an active point $i$ as the sum of three weighted forces

$$\dot{\mathbf{x}}_i = w_{\text{track}} \mathbf{F}_{\text{track},i} + w_{\text{repel}} \mathbf{F}_{\text{repel},i} + w_{\text{bound}} \mathbf{F}_{\text{bound},i}. \quad (5)$$

The scalar weights $w_{\text{track}}$, $w_{\text{repel}}$, and $w_{\text{bound}}$ balance trajectory tracking with self-correcting spatial regularization. Specifically, the tracking force $\mathbf{F}_{\text{track},i}$ is modeled as a linear spring pulling the particle toward its predicted location $\hat{\mathbf{x}}_i$. To maintain a uniform distribution, the repelling force $\mathbf{F}_{\text{repel},i}$ applies pairwise repulsion if the separation between neighboring points drops below a minimum threshold $r_{\text{repel}}$. Simultaneously, the boundary force $\mathbf{F}_{\text{bound},i}$ confines points within the region of interest by exerting an inward force proportional to the gradient of the signed distance field (SDF), $\mathcal{D}^{(t)}$, of the segmentation mask $\mathcal{M}^{(t)}$. Crucially, treating the repelling and boundary forces as soft constraints rather than strict barrier functions permits temporary violations, such as the artificial compression of tightly packed points by a rapidly shrinking mask, which are subsequently resolved during filtering.

**Detection and Removal of Unhealthy Points.** Following the force field, we detect and remove invalid points, specifically outliers and points within overcrowded regions.

A point is classified as an outlier if its SDF value exceeds a margin distance $d_{\text{outlier}}$:

$$\mathcal{P}_{\text{outlier}} = \{ i \in \mathcal{P} \mid \mathcal{D}(\mathbf{x_i}) > d_{\text{outlier}} \}. \quad (6)$$

Overcrowding occurs wherever tracked points violate a minimum safety distance, $d_{\text{crowd}} = l_\rho \alpha_{\text{crowd}}$, where $\alpha_{\text{crowd}}$ is a tunable relaxation constant. To eliminate these violations while retaining the maximum number of tracked points, we formulate the filtering task as a maximum independent set problem on an undirected graph $G_{\text{o}} = (V_{\text{o}}, E_{\text{o}})$. The nodes $V_{\text{o}}$ represent the active points $\mathcal{P}$, and the edges $E_{\text{o}}$ connect points that breach the distance constraint:

$$E = \left\{ \{i, j\} \subset V \mid i \neq j, \ \|\mathbf{x_i} - \mathbf{x_j}\| < d_{\text{crowd}} \right\}. \quad (7)$$

As finding the maximum independent set is NP-hard [36], we approximate the optimal solution using a greedy approach. We iteratively remove the node with

**Algorithm 1:** Trajectory Extraction

**Input :** Video $V$, Video segmentation $\mathcal{M}$, Horizon $h$, Max active query points $\hat{K}_{\max}$, Max segmentation area $A_{\max}$
**Output :** Set of trajectory segments $\mathcal{S}$

```
1  S ← ∅                                              ▷ Initialize global trajectory set
2  P^(1), X^(1) ← KMeans(K̂_max, M^(1))               ▷ Initialize active points via CVT
   ▷ Per-horizon: Track ahead and inject query points
3  for t ← 1 to F − h step h do
4      X̂ ← ψ_T(X^(t), V[t : t + h])                  ▷ Predict raw trajectories
       ▷ Per-frame: physics and removal of unhealthy query points
5      for τ ← t to t + h − 1 do
6          X^(τ+1) ← IntegrateKinematics(X^(τ), X̂^(τ+1), M^(τ))
           ▷ Identify and remove unhealthy points
7          P_crowd ← getOvercrowded(X^(τ+1))
8          P_outlier ← getOutliers(X^(τ+1), M^(τ+1))
9          P_culled ← P_crowd ∪ P_outlier
           ▷ Update indices and filter positions
10         P^(τ+1) ← P^(τ) \ P_culled
11         X^(τ+1) ← FilterPositions(P^(τ+1), X^(τ+1))
           ▷ Add terminated segments to collection
12         S ← finalizeSegments(S, P_culled, X^(τ))
       ▷ Balance Point Density for the next horizon
13     K̂^(t+h) ← ⌈ (K̂_max / A_max) A^(t+h) ⌉
14     if |P^(t+h)| < K̂^(t+h) then
15         K_deficit ← K̂^(t+h) − |P^(t+h)|
16         P^(t+h), X^(t+h) ← InjectQueryPoints(X^(t+h), M^(t+h), K_deficit)
17 S ← finalizeSegments(S, P^(F), X^(F))              ▷ Add remaining active segments
18 return S
```

the highest degree, breaking ties randomly, until the graph becomes fully disconnected. The final set of removed nodes is denoted as $\mathcal{P}_{\text{crowd}}$.

**Tracking Point Insertion.** The emergence of previously occluded surfaces on the main subject, as well as tracking errors, can leave large areas devoid of tracked points. To mitigate this, we periodically insert new points into these low-coverage regions every $h$ frames.To identify candidate insertion areas, we compute a discrete minimum distance map $\mathcal{C}$, representing the shortest distance from each pixel in the segmentation mask to the nearest tracked point. We extract a candidate set $\mathcal{C}_{\text{cov}}$ by filtering out pixels whose distance falls below a coverage threshold $d_{\text{cov}}$

$$\mathcal{C}_{\text{cov}} = \{\mathbf{q} \in \mathcal{M} \mid \mathcal{C}(q) > d_{\text{cov}}\}. \tag{8}$$

Here, $d_{\text{cov}} = l_\rho \alpha_{\text{cov}}$, where $\alpha_{\text{cov}}$ serves as a tunable relaxation coefficient. We populate these uncovered regions by sampling $K_{\text{deficit}} = \hat{K} - |\mathcal{P}|$ points from $\mathcal{C}_{\text{cov}}$ using furthest point sampling (FPS). To maintain adequate spatial distribution, the insertion process terminates early if the distance between any newly sampled point and an existing point drops below $d_{\text{cov}}$.

Because FPS tends to place samples near or directly on the boundaries of the segmentation mask, often leaving large gaps between new and existing setpoints, we introduce a subsequent relaxation step. Specifically, we apply a modified K-means algorithm across the segmentation mask, initializing the centroids with both the new and old points. We then run K-means, but keep the centroids corresponding to preexisting points fixed. Through this, the newly added points are naturally drawn away from the boundaries to evenly fill the gaps.

## Show Planning

The point tracker yields a set of trajectory segments, $\mathcal{S}$. Next, the show planning stage translates these segments into individual setpoint trajectories, defining the complete flight path for each drone from takeoff through the show to landing. The subsequent drone allocation stage has a threefold objective. First, it preprocesses the segments in $\mathcal{S}$ into 3D waypoints for the safety filter, ensuring they are kinematically feasible and mostly collision-free. Second, it assigns these segments to specific drones. Finally, it assembles a continuous trajectory for each drone by merging segment, transition, takeoff, and landing maneuvers.

**Preprocessing.** The initial trajectory segments in $\mathcal{S}$ are defined within the video's 2D image space and time domain. Because these raw segments lack smoothness and contain short segments that complicate downstream safety filtering, we apply a sequential processing pipeline.

First, we filter out noise by discarding segments shorter than five frames. To obtain continuous representations for the remaining valid segments, we fit a cubic B-spline to each. These splines are generated by solving a regularized least-squares problem that penalizes acceleration. The smoothing hyperparameter is selected automatically using the Generalized Cross Validation criterion [37], yielding a set of twice-continuously differentiable splines, $\mathbf{P}$.

Next, we uniformly scale the trajectories spatially to mitigate collisions. To prevent smoothing artifacts from causing too large formations, we employ a percentile-based approach: we sample the minimum pairwise distances between active drones across a dense time grid, then compute a spatial scaling factor such that a low percentile of these distances is lower than the drones' defined safety distance. The safety filter subsequently resolves these remaining collisions.

To ensure the choreography is physically executable, we then rescale the timeline. By sampling the maximum velocities and accelerations across $\mathbf{P}$, we compute a temporal scaling factor that bounds the global extrema strictly below the hardware limits ($v_{\max}$ and $a_{\max}$), adjusted by tunable buffer coefficients. Lowering these buffers preserves kinematic headroom for subsequent transition maneuvers.

Finally, we project the spatially and temporally scaled trajectories from the 2D image plane onto a designated 3D physical plane in the sky.

**Drone Allocation.** To synthesize a continuous choreography from disjoint trajectory segments, we must assign these segments to individual drones and generate the necessary flight transitions. We begin by directly allocating all trivial segments, i.e., those spanning the entire video duration, to individual drones. All remaining assignments are determined by solving a minimum-cost flow problem over a directed graph $G = (V, E)$, where each unit of flow corresponds to a single routed drone.

The network graph $G = (V, E)$ models the drone routing problem. The vertex set $V$ comprises four subsets of nodes.

- **Activation nodes** $\mathcal{A}$**:** Each node $i \in \mathcal{A}$ represents the start of a trajectory segment at time $t_{i,\text{start}}$ with a demand of $b(i) = 1$.
- **Deactivation nodes** $\mathcal{D}$**:** Each node $i \in \mathcal{D}$ represents the end of a trajectory segment at time $t_{i,\text{end}}$ with a demand of $b(i) = -1$.
- **Source node** $s_g$**:** Represents the pool of unassigned, available drones. Its balance is $b(s_g) = -K_{\text{avail}}$, where $K_{\text{avail}}$ is the total number of drones minus the ones assigned to the trivial segments.
- **Sink node** $t_g$**:** Represents the end-of-show retirement state for the drones. Its balance is $b(t_g) = K_{\text{avail}} + |\mathcal{D}| - |\mathcal{A}|$.

The edge set $E$ defines feasible drone transitions, i.e., transitions that drones could fly in principle. Each edge $(i, j) \in E$ is assigned a flow capacity $u_{ij}$ and a cost $c_{ij}$. We have the following edges:

- **Transition edges** $\mathcal{D} \to \mathcal{A}$**:** Connect a deactivation node $i \in \mathcal{D}$ to a subsequent activation node $j \in \mathcal{A}$. Flow along such an edge represents the reassignment of a drone to a new segment. It exists only if the transition is temporally feasible:

$$t_{ij,\text{req}} \leq t_{j,\text{start}} - t_{i,\text{end}}, \tag{9}$$

  where $t_{ij,\text{req}}$ is a conservative estimate of the transition duration. The cost $c_{ij}$ penalizes both the transition distance and the kinematic misalignment between the transition direction, $\mathbf{d}_{ij} = \frac{\mathbf{p}_j(t_{j,\text{start}}) - \mathbf{p}_i(t_{i,\text{end}})}{\|\mathbf{p}_j(t_{j,\text{start}}) - \mathbf{p}_i(t_{i,\text{end}})\|}$, and the boundary velocities:

$$c_{ij,\text{dist}} = \|\mathbf{p}_j(tj, \text{start}) - \mathbf{p}_i(ti, \text{end})\|, \tag{10}$$

$$c_{i,\text{align}} = \|\dot{\mathbf{p}}_i(t_{i,\text{end}})\| \left(1 - \frac{\dot{\mathbf{p}}_i(t_{i,\text{end}}) \cdot \mathbf{d}ij}{\|\dot{\mathbf{p}}_i(ti, \text{end})\|}\right), \tag{11}$$

$$c_{j,\text{align}} = \|\dot{\mathbf{p}}_j(t_{j,\text{start}})\| \left(1 - \frac{\dot{\mathbf{p}}_j(t_{j,\text{start}}) \cdot \mathbf{d}ij}{\|\dot{\mathbf{p}}_j(tj, \text{start})\|}\right), \tag{12}$$

$$c_{ij} = c_{ij,\text{dist}} + w_{\text{align}} \left(c_{i,\text{align}} + c_{j,\text{align}}\right), \tag{13}$$

  where $w_{\text{align}} > 0$ is a weighting factor.
- **Takeoff edges** $s_g \to \mathcal{A}$**:** Connect the source to an activation $j \in \mathcal{A}$, representing the deployment of a fresh drone from the landing pad. We apply a large constant penalty $M$ to minimize new drone allocations.
- **Landing edges** $\mathcal{D} \to t_g$**:** Connect deactivations $i \in \mathcal{D}$ to the sink. Flow represents a drone returning to the landing pad without reassignment. This edge incurs zero cost, as every active drone must eventually land.
- **Bypass edge** $s_g \to t_g$**:** Connects the source directly to the sink. Flow represents drones that remain grounded. This edge has zero cost.

All edges have a unit capacity ($u_{ij} = 1$), ensuring only one drone flies a given transition, except the bypass edge, whose capacity is infinite.

We formulate the assignment as a Minimum Cost Network Flow (MCNF) problem. We seek the optimal flow $f : E \to \mathbb{N}_0$ that minimizes the total cost while satisfying capacity limits and balancing node demands:

$$\min_f \sum_{(i,j)\in E} c_{ij} f(i,j) \tag{14}$$

$$\text{s.t. } 0 \leq f(i,j) \leq u_{ij}, \quad \forall (i,j) \in E,$$

$$\sum_{(i,j)\in\delta^+(i)} f(i,j) - \sum_{(j,i)\in\delta^-(i)} f(j,i) = b(i), \quad \forall i \in V,$$

where $\delta^+(i)$ and $\delta^-(i)$ denote the sets of outgoing and incoming edges at node $i$, respectively.

In edge cases, the planner might fail if the number of simultaneously active trajectory segments ($\hat{K}_{\max}$) is too large. To mitigate this, we use the following strategy. We start with setting $\hat{K}_{\max} = K$ inside the point tracker and run tracker and planner. In case the planner cannot find a solution, we temporarily relax $K_{\text{avail}}$ to a sufficiently large value, $K^*_{\text{avail}}$, until feasibility is achieved. The resulting drone deficit is calculated as $K_{\text{miss}} = K^*_{\text{avail}} - K_{\text{avail}} - f(s_g, t_g)$. The tracking process is then rerun by reducing $\hat{K}_{\max}$ by this value until the planner finds a solution.

Once the optimal flow $f^*$ is found, final assignments are extracted by iteratively tracing single units of flow through the graph. Reaching an activation node logically transitions the path to the corresponding deactivation node, dictating that the assigned drone executes that specific trajectory segment.

**Trajectory Assembly** Following segment assignment, each drone $k$ receives an ordered sequence of trajectory segments and corresponding transition durations. To construct executable trajectories, we interleave smooth transition maneuvers between these segments, bounded by defined takeoff and landing phases. The resulting merged trajectory, $\Gamma_k(t)$, is constrained to maintain $C^2$ continuity across all boundaries. We generate these paths using a three-step assembly algorithm:

1. **Takeoff and Landing Routing:** We assign drones to specific ground start and end positions via the Hungarian algorithm. This establishes an initial routing configuration that minimizes collision risks during deployment and recovery [13, 17].
2. **Greedy Transition Insertion:** We iterate through all required transitions, prioritizing them by duration. Each transition is formulated and solved as a boundary value problem (BVP) in the image plane to guarantee $C^2$ continuity. If inter-drone collisions are detected between a new transition and existing segments, we iteratively apply scaling orthogonal bump profiles to the trajectory until the spatial conflict clears, or the iteration limit is reached.
3. **Global Time Reparameterization:** Because spatial collision-avoidance adjustments may induce kinematic violations (e.g., exceeding velocity or acceleration limits), all merged trajectories undergo a final, uniform time reparameterization to ensure dynamic feasibility.

## Safety Filter

As the final stage of our proposed pipeline, similar to existing frameworks [13, 14, 15, 16, 17], we employ a safety filter to resolve collisions. More specifically, we adopt the approach by Schuck et al. [17] and use the state-of-the-art motion planner AxSwarm [20, 38] as a safety filter. It ensures the kinematic feasibility and collision-free execution of the final trajectories. Specifically, AxSwarm generates trajectory via solving a distributed model predictive control (MPC) problem. In this context, "distributed" refers to the decoupled optimization of each drone's trajectory to facilitate efficient parallel computation [38], rather than the physical distribution of computational resources onboard the agents.

The underlying dynamics are represented by a discrete-time linear state-space model [17]

$$\mathbf{x}_{k,i+1} = A\mathbf{x}_{k,i} + B\mathbf{u}_{k,i}, \tag{15}$$

where $i$ denotes the time step index. The state vector for drone $k$ at time $t_i$ is defined as $\mathbf{x}_{k,i} = [q_k(t_i)^T, \dot{q}_k(t_i)^T]^T$, and $\mathbf{u}_{k,i} \in \mathbb{R}^6$ represents the high-level control input, comprising the reference position and velocity. The system matrices $A$ and $B$ govern the evolution of the drone's spatial state under an underlying low-level Mellinger controller [39] that tracks the reference input $\mathbf{u}_{k,i}$.

For each time step $t_i \in \{0, 1, \ldots, T\}$ throughout the sequence, AxSwarm computes the optimal control inputs over a finite prediction horizon of $N$ steps. This optimization aims to accurately track the reference trajectories $\Gamma_k$ while ensuring the resulting motion remains smooth, dynamically feasible, and safe. Dropping the drone index $k$ for notational brevity, the corresponding optimization problem for an individual drone is formulated as follows [17, 20]:

$$\min_{\mathbf{u}_{\cdot|i}} \sum_{n=1}^{N} \left( w_{\text{pos}} \|\ddot{\mathbf{q}}_{n+1|i}\|^2 + w_{\text{ctr}} \|\ddot{\mathbf{u}}_{n|i}\|^2 \right) \tag{16a}$$

$$\text{s.t. } \mathbf{x}_{n+1|i} = A\mathbf{x}_{n|i} + B\mathbf{u}_{n|i}, \forall n \in \{1, \ldots, N\} \tag{16b}$$

$$\mathbf{q}_{s|i} = \Gamma(t_s), \forall s \in \mathbb{S}_i \tag{16c}$$

$$\mathbf{x}_{i|i} = \mathbf{x}_i, \tag{16d}$$

$$\mathbf{u}^{(d)}_{i|i} = \mathbf{u}^{(d)}_{i|i-1}, \forall d \in \{0, 1, 2\} \tag{16e}$$

$$\|\dot{\mathbf{q}}_{n|i}\|^2 \leq v^2_{\max}, \forall n \in \{1, \ldots, N\} \tag{16f}$$

$$f^2_{\min} \leq \|\ddot{\mathbf{q}}_{n|i} + \mathbf{g}\|^2 \leq f^2_{\max}, \forall n \in \{1, \ldots, N\} \tag{16g}$$

$$\|\mathbf{q}_{n|i} - \mathbf{q}_{j,n|i-1}\|^2 \geq \alpha_{\text{dist},2} d_{\text{dist}}, \forall j \in \mathbb{J}, \forall n \in \{1, \ldots, N\} \tag{16h}$$

where the subscript $n|i$ means a prediction $n$ steps ahead made at time point $i$.

To enforce trajectory smoothness, the optimization objective minimizes the squared accelerations of both the predicted position and the control input. Although the control input $\mathbf{u}$ is sampled discretely in this formulation, AxSwarm continuously parameterizes it using a 10th-order Bernstein polynomial over the prediction horizon.

The optimization is subject to several kinematic and operational constraints. Equation (16c) ensures that the generated trajectory accurately tracks the reference path $\Gamma$ at a predefined set of sampling points $\mathbb{S}_i$, decoupled from the MPC time steps. Additionally, Equation (16e) guarantees the continuity of the control input across consecutive time steps. Equation (16f) limits the velocity of each drone. For efficient distributed collision checking, AxSwarm only evaluates potential collisions against the previous time step's predictions for a local subset $\mathbb{J}$ of neighboring drones. Furthermore, the collision constraint scales the nominal threshold distance $d_{\text{dist}}$ by a safety factor $\alpha_{\text{dist},2}$. Selecting $\alpha_{\text{dist},2} \leq 1$ introduces a buffer to account for minor numerical constraint violations by the solver.

In practice, this optimization problem is solved iteratively at a fixed frequency $f_{\text{mpc}}$. Each iteration yields the optimal control sequence $\mathbf{u}^*_{i:N|i}$ and the corresponding predicted state trajectory $\mathbf{x}^*_{i:N|i}$ over the full horizon. Following standard receding-horizon control principles, we retain only the predicted state for the immediate next time step. The final sequence of safe, dynamically feasible waypoints is then assembled by combining the waypoints

$$\Gamma_{\text{safe}} = \left\{ \mathbf{q}^*_{1|0}, \mathbf{q}^*_{1|1}, \ldots, \mathbf{q}^*_{1|T} \right\}, \tag{17}$$

which are then sent to the drones.

## Acknowledgments

We thank the Bitcraze team for their assistance with the hardware experiments and filming. Special thanks to Aris Morsink Paloumpas for setting up the testbed, and to both Aris and Barbara Rousselot for their help with filming.

This work has been supported by the German Federal Ministry of Research, Technology and Space (BMFTR) under the Robotics Institute Germany (RIG). We acknowledge EuroHPC Joint Undertaking for awarding us access to MareNostrum5 at BSC, Spain.